\documentclass[runningheads]{llncs}
\usepackage[T1]{fontenc}
\usepackage{graphicx}
\usepackage{multicol,multirow}
\usepackage{amsmath}
\usepackage{wrapfig}
\usepackage{booktabs}       
\usepackage{amsfonts}       
\usepackage[table,xcdraw]{xcolor}         
\usepackage{setspace} 
\usepackage{algorithm}
\usepackage{algorithmic}
\usepackage{tabularx} 
\usepackage{array} 
\begin{document}
\title{Q-YOLO: Efficient Inference for Real-time Object Detection \thanks{$\dag$ Equal contribution. $\star$ Corresponding author.}}
%
\author{
Mingze Wang $^{\dag}$ \and
Huixin Sun $^{\dag}$\and
Jun Shi $^{\dag}$ \and
Xuhui Liu \and
Baochang Zhang \and 
Xianbin Cao$^{\star}$
}
\institute{Beihang University, Beijing, China \\
\email{\{wmz20000729,sunhuixin,ShiJun2020,1332671326,bczhang,xbcao\}@buaa.edu.cn}}
\maketitle              
\begin{abstract}
Real-time object detection plays a vital role in various computer vision applications. However, deploying real-time object detectors on resource-constrained platforms poses challenges due to high computational and memory requirements. This paper describes a low-bit quantization method to build  a highly efficient one-stage detector, dubbed as  Q-YOLO, which can effectively address the performance degradation problem caused by activation distribution imbalance in traditional quantized YOLO models. Q-YOLO introduces a fully end-to-end Post-Training Quantization (PTQ) pipeline with a well-designed Unilateral Histogram-based (UH) activation quantization scheme, which determines the maximum truncation values through histogram analysis by minimizing the Mean Squared Error (MSE) quantization errors. Extensive experiments on the COCO dataset demonstrate the effectiveness of Q-YOLO, outperforming other PTQ methods while achieving a more favorable balance between accuracy and computational cost. This research contributes to advancing the efficient deployment of object detection models on resource-limited edge devices, enabling real-time detection with reduced computational and memory overhead.
\end{abstract}
\section{Introduction}
Real-time object detection is a crucial component in various computer vision applications, such as multi-object tracking~\cite{zhang2021fairmot,zhang2022bytetrack}, autonomous driving~\cite{li2019gs3d,feng2020deep}, and robotics~\cite{karaoguz2019object,paul2021object}. The development of real-time object detectors, particularly YOLO-based detectors, has yielded remarkable performance in terms of accuracy and speed. For example, YOLOv7-E6~\cite{wang2023yolov7} object detector achieves 55.9\% mAP on COCO {\tt 2017}, outperforming both transformer-based detector SWINL Cascade-Mask R-CNN~\cite{liu2021swin,cai2018cascade} and convolutional based detector ConvNeXt-XL Cascade-Mask R-CNN~\cite{woo2023convnext,cai2018cascade} in both speed and accuracy. Despite their success, the computational cost during inference remains a challenge for real-time object detectors on resource-limited edge devices, such as mobile CPUs or GPUs, limiting their practical usage.
\begin{figure}[t]
\begin{center}
\centerline{\includegraphics[width=1\linewidth]{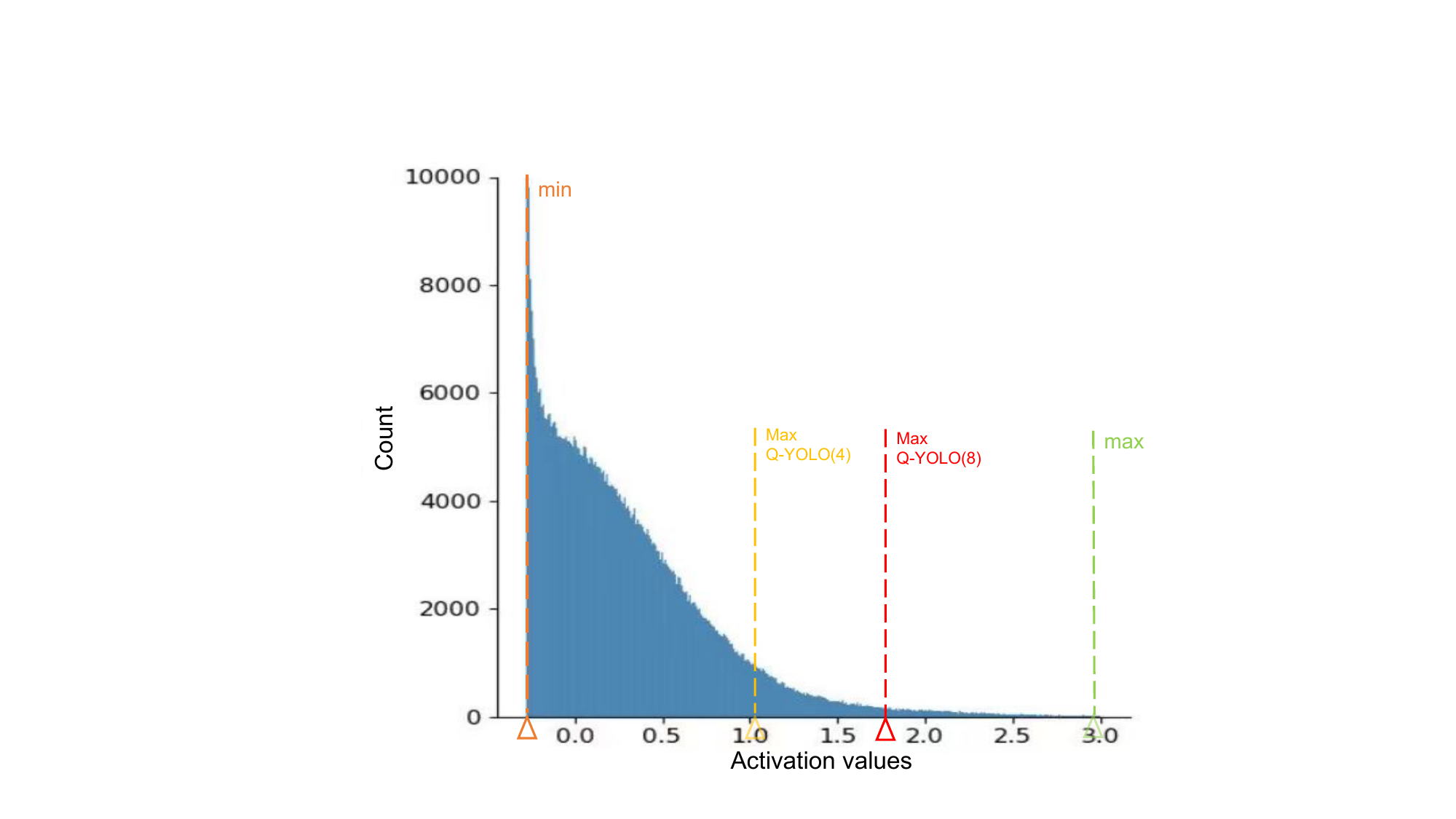}}
\caption{Activation value distribution histogram (with 2048 bins) of the model.21.conv layer in YOLOv5s. The occurrence of values between 0 and -0.2785 is extremely high, while the frequency of values above zero decreases significantly, reveals an imbalanced pattern. {\color{orange}\textbf{min}} denotes the fixed minimum truncation value, while {\color{green}\textbf{max}} represents the maximum truncation value following the min-max principle. {\color{red}\textbf{Max Q-YOLO(8)}} refers to the maximum truncation value when using the Q-YOLO quantization model at 8-bit, and {\color{yellow}\textbf{Max Q-YOLO(4)}} indicates the maximum truncation value when applying the Q-YOLO quantization model at 4-bit.} 
\label{fig:histogram}
\end{center}
\vspace{-10mm}
\end{figure}
Substantial efforts on network compression have been made towards efficient online inference~\cite{xu2022recurrent,romero2014fitnets,qin2020forward,denil2013predicting}. Methods include enhancing network designs~\cite{howard2017mobilenets,wu2018shift,zhang2018shufflenet}, conducting network search~\cite{zoph2018learning}, network pruning~\cite{han2015compressing,guo2016dynamic}, and network quantization~\cite{li2022q}. Quantization, in particular, has gained significant popularity for deployment on AI chips by representing a network using low-bit formats. There are two prevailing quantization methods, Quantization-Aware Training (QAT)~\cite{li2022q,xu2023q} and Post-Training Quantization (PTQ)~\cite{lin2021fq}. Although QAT generally achieves better results than PTQ, it requires training and optimization of all model parameters during the quantization process. The need for pretraining data and significant GPU resources makes QAT challenging to execute. On the other hand, PTQ is a more efficient approach for quantizing real-time object detectors.

To examine low-bit quantization for real-time object detection, we first establish a PTQ baseline using YOLOv5~\cite{yolov5}, a state-of-the-art object detector. Through empirical analysis on the COCO {\tt 2017} dataset, we observe notable performance degradation after quantization, as indicated in Table~\ref{exp_main}. For example, a 4-bit quantized YOLOv5s employing Percentile achieves only 7.0\% mAP, resulting in a performance gap of 30.4\% compared to the original real-valued model. We find the performance drop of quantized YOLOs can be attributed to the activation distribution imbalance. As shown in Fig.~\ref{fig:histogram}, we observe high concentration of values close to the lower bound and the significant decrease in occurrences above zero. When employing fixed truncation values such as MinMax, representing activation values with extremely low probabilities would consume a considerable number of bits within the limited integer bit width, resulting in further loss of information.

\begin{figure}[t]
\begin{center}
\centerline{\includegraphics[width=1\linewidth]{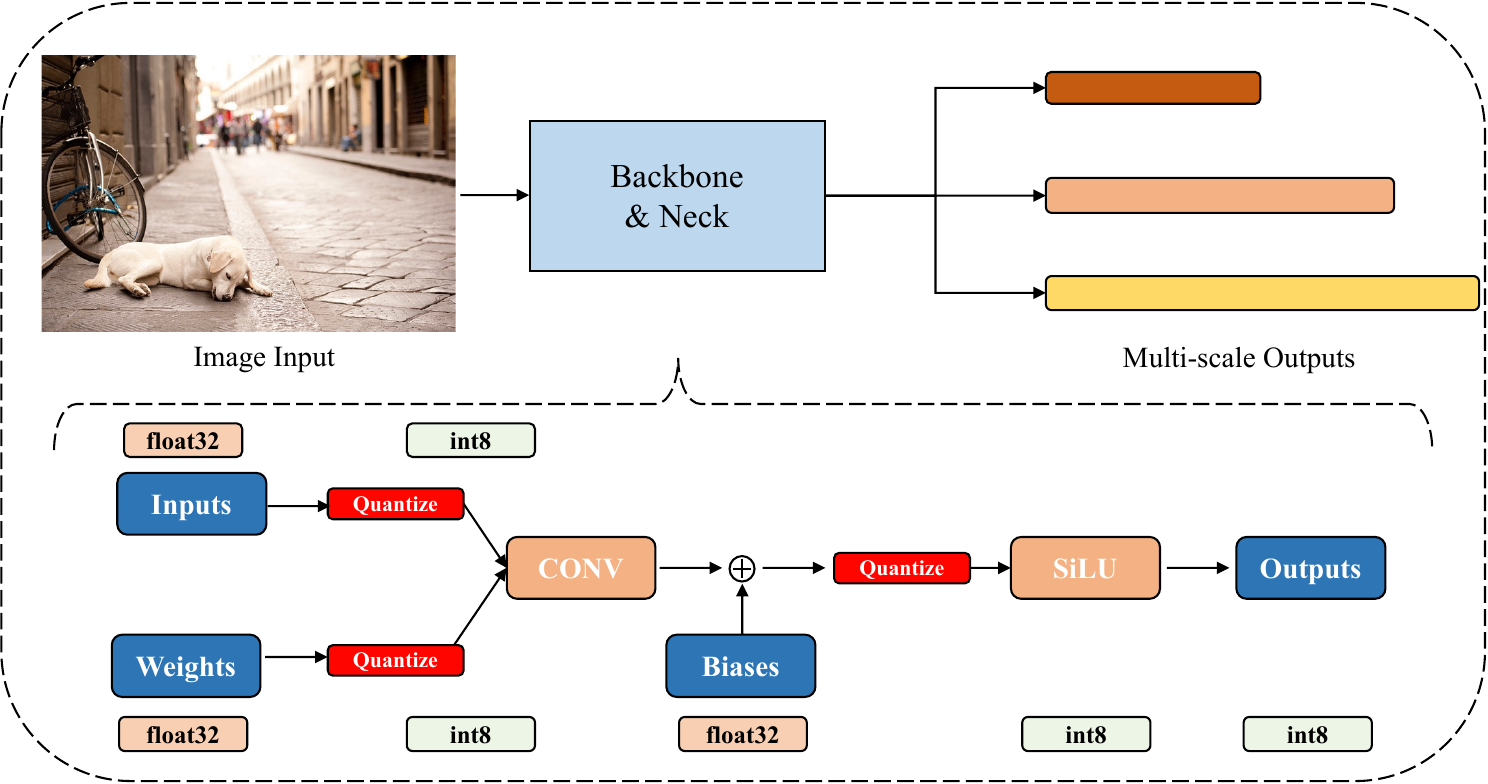}}
\caption{Architecture of Q-YOLO.} 
\label{fig:arch}
\end{center}
\vspace{-10mm}
\end{figure}

In light of the above issue, we introduce Q-YOLO, a fully end-to-end PTQ quantization architecture for real-time object detection, as depicted in Fig.~\ref{fig:arch}. Q-YOLO quantizes the backbone, neck, and head modules of YOLO models, while employing standard MinMax quantization for weights. To tackle the problem of activation distribution imbalance, we introduce a novel approach called Unilateral Histogram-based (UH) activation quantization. UH iteratively determines the maximum truncation value that minimizes the quantization error through histograms. This technique significantly reduces calibration time and effectively addresses the discrepancy caused by quantization, optimizing the quantization process to maintain stable activation quantization. By mitigating information loss in activation quantization, our method ensures accurate object detection results, thereby enabling precise and reliable low-bit real-time object detection performance. Our contributions can be summarized as follows:
\begin{enumerate}
\item We introduce  a fully end-to-end PTQ quantization architecture specifically designed for real-time object detection, dubbed as Q-YOLO.
\item A Unilateral Histogram-based (UH) activation quantization method is proposed to  leverage  histogram analysis to find the maximum truncation values, which can effectively  minimize the MSE quantization error.
\item Through extensive experiments on various object detectors, we demonstrate that Q-YOLO outperforms baseline PTQ models by a significant margin. The 8-bit Q-YOLO model applied on YOLOv7 achieves a 3$\times$ acceleration while maintaining performance comparable to its full-precision counterpart on COCO, highlighting its potential as a general solution for quantizing real-time object detectors.
\end{enumerate}

\section{Related Work}
\subsection{Quantization}
Quantized neural networks are based on low-bit weights and activations to accelerate model inference and save memory. The commonly used model quantization methods include  quantization-aware training (QAT)  and  post-training quantization (PTQ). In QAT, Zhang \emph{et al.}~\cite{zhang2021modulated} builds a binarized convolutional neural network based on a projection function and a new updated rule during the backpropagation. Li \emph{et al.}~\cite{li2022q} proposed an information rectification module and distribution-guided distillation to push the bit-width in a quantized vision transformer. TTQ~\cite{zhu2016trained} uses two real-valued scaling coefficients to quantize the weights to ternary values. Zhuang \emph{et al.} \cite{zhuang2018towards} present a low-bit (2-4 bit) quantization scheme using a two-stage approach to alternately quantize the weights and activations, providing an optimal trade-off among memory, efficiency, and performance. In~\cite{jung2019learning}, the quantization intervals are parameterized, and optimal values are obtained by directly minimizing the task loss of the network. ZeroQ~\cite{cai2020zeroq} supports uniform and mixed-precision quantization by optimizing for a distilled dataset which is engineered to match the statistics of the batch normalization across different network layers. \cite{fang2020post} enabled accurate approximation for tensor values that have bell-shaped distributions with long tails and found the entire range by minimizing the quantization error.While QAT often requires high-level expert knowledge and huge GPU resources for training or fine-tuning, especially the large-scale pre-trained model. To reduce the above costs of quantization, PTQ, which is training-free, has received more widespread attention and lots of excellent works arise. MinMax, EMA~\cite{jacob2018quantization} methods are commonly used to compress or reduce the weights of the PTQ model. MinMax normalizes the weights and bias values in the model to a predefined range, such as [-1, 1], to reduce the storage space and increase the inference speed. MSE quantization involves evaluating and adjusting the quantized activation values to minimize the impact of quantization on model performance.

\subsection{Real-time Object Detection}
Deep Learning based object detectors can be generally classified into two categories: two-stage and single-stage object detectors. Two-stage detectors, such as Faster R-CNN~\cite{ren2015faster}, RPN~\cite{lin2017feature}, and Cascade R-CNN~\cite{cai2018cascade}, first generate region proposals and then refine them in a second stage. On the other hand, single-stage object detectors have gained significant popularity in real-time object detection due to their efficiency and effectiveness. These detectors aim to predict object bounding boxes and class labels in a single pass of the neural network, eliminating the need for time-consuming region proposal generation. One of the pioneering single-shot detectors is YOLO~\cite{redmon2016you}, which divides the input image into a grid and assigns bounding boxes and class probabilities to predefined anchor boxes. The subsequent versions, YOLOv2~\cite{redmon2017yolo9000} and YOLOv3~\cite{redmon2018yolov3}, introduced improvements in terms of network architecture and feature extraction, achieving better accuracy without compromising real-time performance. Another influential single-shot detector is SSD~\cite{liu2016ssd}, which employs a series of convolutional layers at different scales to detect objects of various sizes. By using feature maps at multiple resolutions, SSD achieves high accuracy while maintaining real-time performance. Variants of SSD, such as MobileNet-SSD~\cite{howard2017mobilenets} and Pelee~\cite{wang2018pelee}, further optimize the architecture to achieve faster inference on resource-constrained devices.

Efficiency is a critical aspect of real-time object detection, especially for deployment on computationally limited platforms. MobileNet\cite{howard2017mobilenets} and its subsequent variants, such as MobileNetV2\cite{sandler2018mobilenetv2} and MobileNetV3~\cite{koonce2021mobilenetv3}, have received significant attention for their lightweight architectures. These networks utilize depth-wise separable convolutions and other techniques to reduce the number of parameters and operations without significant accuracy degradation. ShuffleNet\cite{zhang2018shufflenet} introduces channel shuffling operations to exploit group convolutions, enabling a trade-off between model size and computational cost. ShuffleNetV2\cite{ma2018shufflenet} further improves the efficiency by introducing a more efficient block design and exploring different network scales.

\section{Methodology}
\subsection{Preliminaries}
\subsubsection{Network Quantization Process.} 
We first review the main steps of the Post-Training Quantization (PTQ) process and supply the details. Firstly, the network is either trained or provided as a pre-trained model using full precision and floating-point arithmetic for weights and activations. Subsequently, numerical representations of weights and activations are suitably transformed for quantization. Finally, the fully-quantized network is deployed either on integer arithmetic hardware or simulated on GPUs, enabling efficient inference with reduced memory storage and computational requirements while maintaining reasonable accuracy levels. 
\subsubsection{Uniform Quantization.} Assuming the quantization bit-width is $b$, the quantizer $\textrm{Q}(\mathbf{x}|b)$ can be formulated as a function that maps a floating-point number $\mathbf{x}\in \mathbb{R}$ to the nearest quantization bin:
\begin{equation}
    \textrm{Q}(\mathbf{x}|b): \mathbb{R} \rightarrow \hat{\mathbf{x}},
\end{equation}
\begin{equation}
\hat{\mathbf{x}}=
\left\{\begin{aligned}
&\{-\textrm{2}^{b-1},\cdots ,\textrm{2}^{b-1}-\textrm{1}\} &\text{Signed},\\
&\{\textrm{0} \cdots ,\textrm{2}^{b}-\textrm{1}\}  &\text{Unsigned}.
\end{aligned}\right.
\label{eq:signed}
\end{equation}
There are various quantizer $\textrm{Q}(\mathbf{x}|b)$, where uniform~\cite{jacob2018quantization} are typically used. Uniform quantization is well supported on most hardware platforms. Its unsigned quantizer $\textrm{Q}(\mathbf{x}|b)$ can be defined as:
\begin{equation}
    \textrm{Q}(\mathbf{x}|b)=\operatorname{clip}(\lfloor\frac{\mathbf{x}}{s_\mathbf{x}}\rceil+zp_\mathbf{x}, \textrm{0}, \textrm{2}^{b}-\textrm{1}),
\label{eq:quant}
\end{equation}
where $s_\mathbf{x}$~(scale) and $zp_\mathbf{x}$~(zero-point) are quantization parameters. In Eq.~\ref{eq:factor}, $u$ (upper) and $l$ (lower) define the quantization grid limits. 
\begin{align}
    s_\mathbf{x}=&\frac{u-l}{\textrm{2}^b-\textrm{1}},zp_\mathbf{x}=\operatorname{clip}(\lfloor-\frac{l}{s}\rceil, \textrm{0}, \textrm{2}^{b}-\textrm{1}).
\label{eq:factor}
\end{align}
The dequantization process can be formulated as: 
\begin{equation}
    \Tilde{\mathbf{x}}=(\hat{\mathbf{x}}-zp_\mathbf{x}) \times s_\mathbf{x}.
\end{equation}
\subsection{Quantization Range Setting} 
\label{quantization_range_setting}
Quantization range setting is the process of establishing the upper and lower clipping thresholds, denoted as $u$ and $l$ respectively, of the quantization grid. The crucial trade-off in range setting lies in the balance between two types of errors: clipping error and rounding error. Clipping error arises when data is truncated to fit within the predefined grid limits, as described in Eq.\ref{eq:factor}. Such truncation leads to information loss and a decrease in precision in the resulting quantized representation. On the other hand, rounding error occurs due to the imprecision introduced during the rounding operation, as described in Eq.\ref{eq:quant}. This error can accumulate over time and have an impact on the overall accuracy of the quantized representation. The following methods provide different trade-offs between the two quantities.
\subsubsection{MinMax.} In the experiments, we use the MinMax method for weight quantization, where clipping thresholds $l_\mathbf{x}$ and $u_\mathbf{x}$ are formulated as:
\begin{align}
l_\mathbf{x}=&\min(\mathbf{x}), u_\mathbf{x}=\max(\mathbf{x}).
\label{eq:minmax}
\end{align}
This leads to no clipping error. However, this approach is sensitive to outliers as strong outliers may cause excessive rounding errors.
\subsubsection{Mean Squared Error (MSE).} One way to mitigate the problem of large outliers is by employing MSE-based range setting. In this method, we determine $l_\mathbf{x}$ and $u_\mathbf{x}$ that minimize the mean squared error (MSE) between the original and quantized tensor:

\begin{equation}
\underset{l_\mathbf{x}, u_\mathbf{x}}{\text{arg min}} \; \text{MSE}(\mathbf{x}, \mathbf{Q}_{l_\mathbf{x}, u_\mathbf{x}}),
\end{equation}
where $\mathbf{x}$ represents the original tensor and $\mathbf{Q}_{l_\mathbf{x}, u_\mathbf{x}}$ denotes the quantized tensor produced using the determined clipping thresholds $l_\mathbf{x}$ and $u_\mathbf{x}$. The optimization problem is commonly solved using grid search, golden section method or analytical approximations with closed-form solution.

\subsection{Unilateral Histogram-based (UH) Activation Quantization}
To address the issue of activation value imbalance, we propose a new approach called Unilateral Histogram-based (UH) activation quantization. We first provide an empirical study of the activation values after forward propagation through the calibration dataset. As depicted in Figure~\ref{fig:histogram}, we observe a concentrated distribution of values near the lower bound, accompanied by a noticeable decrease in occurrences above zero. Further analysis of the activation values reveals that the empirical value of -0.2785 serves as the lower bound. This phenomenon can be attributed to the frequent utilization of the Swish (SILU) activation function in the YOLO series.

\begin{algorithm}[ht]
  \caption{Unilateral Histogram-based (UH) Activation Quantization}
  \label{alg:uh_algorithm}
  \begin{algorithmic}[1]
    \STATE \textbf{Input:} FP32 Histogram $H$ with 2048 bins
        \FOR{$i$ \textbf{in} range(128, 2048)}
                \STATE Reference distribution $P \leftarrow H[0:i]$
                \STATE Outliers count $c \leftarrow \sum_{j=i}^{2047} H[j]$
                \STATE $P[i-1] \leftarrow P[i-1] + c$
                \STATE $P \leftarrow \frac{P}{\sum_{j}(P[j])}$
                
                \STATE Candidate distribution $C \leftarrow \text{Quantize } H[0:i] \text{ into 128 levels}$
                \STATE Expand $C$ to have $i$ bins
                
                \STATE $Q \leftarrow \frac{C}{\sum_{j}(C[j])}$
                
                \STATE $MSE[i] \leftarrow \text{Mean Squared Error}(P, Q)$
        \ENDFOR
    \STATE \textbf{Output:} Index $m$ for which $MSE[m]$ is minimal.
  \end{algorithmic}
\end{algorithm}

Based on the empirical evidence, we introduce an asymmetric quantization approach called Unilateral Histogram-based (UH) activation quantization. In UH, we iteratively determine the maximum truncation value that minimizes the quantization error, while keeping the minimum truncation value fixed at -0.2785, as illustrated in the following:
\begin{align}
u_\mathbf{x}= \underset{l_\mathbf{x}, u_\mathbf{x}}{\text{arg min}} \; \text{MSE}(\mathbf{x}, \mathbf{Q}_{l_\mathbf{x},  
  u_\mathbf{x}}), 
l_\mathbf{x}=-0.2785.
\label{eq:uh}
\end{align}

To evaluate the quantization error during the search for the maximum truncation value, we utilize the fp32 floating-point numbers derived from the center values of the gathered 2048 bins, as introduces in Algorithm~\ref{alg:uh_algorithm}. These numbers are successively quantized, considering the current maximum truncation value under consideration. Through this iterative process, we identify the optimal truncation range. The UH activation quantization method offers two key advantages. Firstly, it significantly reduces calibration time. Secondly, it ensures stable activation quantization by allowing a larger set of integers to represent the frequently occurring activation values between 0 and -0.2785, thereby improving quantization accuracy.


%
\section{Experiments}
In order to assess the performance of the proposed Q-YOLO detectors, we conducted a comprehensive series of experiments on the widely recognized COCO {\tt 2017} \cite{lin2014microsoft} detection benchmark. As one of the most popular object detection datasets, COCO {\tt 2017} \cite{lin2014microsoft} has become instrumental in benchmarking state-of-the-art object detectors, thanks to its rich annotations and challenging scenarios. Throughout our experimental analysis, we employed standard COCO metrics on the bounding box detection task to evaluate the efficacy of our approach.

\begin{table*}[!t]
\centering
\caption{A comparison of various quantization methods applied to YOLOv5s~\cite{yolov5}, YOLOv5m~\cite{yolov5}, YOLOv7~\cite{wang2023yolov7} and YOLOv7x\cite{wang2023yolov7}, which have an increasing number of parameters, on the COCO {\tt val2017} dataset~\cite{lin2014microsoft}. The term Bits (W-A) represents the bit-width of weights and activations. The best results are displayed in bold.}
\small
\begin{tabular}{ccccccccccc}
\hline
Models                   & Method      & Bits                 & Size$_{\rm (MB)}$  & OPs$_{\rm (G)}$  & AP   & AP$_{50}$  & AP$_{75}$  & AP$_{s}$  & AP$_{m}$  & AP$_{l}$  \\ \hline
\multirow{6}{*}{YOLOv5s~\cite{yolov5}} & Real-valued & 32-32                & 57.6               & 16.5             & 37.4 & 57.1       & 40.1       & 21.6      & 42.3      & 48.9      \\ \cline{2-11} 
                         & MinMax      & \multirow{3}{*}{8-8} & \multirow{3}{*}{14.4} & \multirow{3}{*}{4.23}            & 37.2 & 56.9 & 39.8 & 21.4 & 42.2 & 48.5 \\
                         & Percentile~\cite{li2019fully}  & & &                                                                        & 36.9 & 56.4 & 39.6 & 21.3 & 42.4 & 48.1 \\
                         & \textbf{Q-YOLO}      &      &  &      & \textbf{37.4} & \textbf{56.9} & \textbf{39.8} & \textbf{21.4} & \textbf{42.4} & \textbf{48.8} \\  \cline{2-11} 
                         & Percentile~\cite{li2019fully}  & \multirow{2}{*}{4-4} & \multirow{2}{*}{7.7} & \multirow{2}{*}{2.16}            & 7.0    & 14.2 & 6.3  & 4.1  & 10.7 & 7.9  \\
                         & \textbf{Q-YOLO}      &      & &                & \textbf{14.0}   & \textbf{26.2} & \textbf{13.5} & \textbf{7.9}  & \textbf{17.6} & \textbf{19.0}   \\ \hline
                         
\multirow{6}{*}{YOLOv5m~\cite{yolov5}} & Real-valued & 32-32                & 169.6               & 49.0            & 45.1 & 64.1 & 49   & 28.1 & 50.6 & 57.8 \\ \cline{2-11} 
                         & MinMax      & \multirow{3}{*}{8-8} & \multirow{3}{*}{42.4} & \multirow{3}{*}{12.4}  & 44.9 & 64   & 48.9 & 27.8 & 50.5 & 57.4 \\
                         & Percentile~\cite{li2019fully}  & & &                      & 44.6 & 63.5 & 48.4 & 28.4 & 50.4 & 57.8 \\
                         & \textbf{Q-YOLO} & & & & \textbf{45.1} & \textbf{64.1} & \textbf{48.9} & \textbf{28} & \textbf{50.6} & \textbf{57.7} \\ \cline{2-11} 
                         & Percentile~\cite{li2019fully}  & \multirow{2}{*}{4-4} & \multirow{2}{*}{21.2} & \multirow{2}{*}{6.33}  & 19.4 & 35.6 & 19.1 & 14.6 & 28.3 & 17.2 \\
                         & \textbf{Q-YOLO} & & & & \textbf{28.8} & \textbf{46} & \textbf{30.5} & \textbf{15.4} & \textbf{33.8} & \textbf{38.7} \\ \hline
                         
\multirow{6}{*}{YOLOv7~\cite{wang2023yolov7}}  & Real-valued & 32-32                & 295.2               & 104.7           & 50.8 & 69.6 & 54.9 & 34.9 & 55.6 & 66.3 \\ \cline{2-11} 
                         & MinMax      & \multirow{3}{*}{8-8} & \multirow{3}{*}{73.8} & \multirow{3}{*}{27.2}  & 50.6 & 69.5 & 54.8 & 34.1 & 55.5 & 65.9 \\
                         & Percentile~\cite{li2019fully}  & & &                      & 50.5 & 69.3 & 54.6 & 34.5 & 55.4 & 66.2 \\
                         & \textbf{Q-YOLO}      & & &                      & \textbf{50.7} & \textbf{69.5} & \textbf{54.8} & \textbf{34.8}   & \textbf{55.5} & \textbf{66.2} \\ \cline{2-11} 
                         & Percentile~\cite{li2019fully}  & \multirow{2}{*}{4-4} & \multirow{2}{*}{36.9} & \multirow{2}{*}{14.1}  & 16.7    & 26.9 & 17.8  & 10.3  & 20.1 & 20.2  \\
                         & \textbf{Q-YOLO}      & & &                      & \textbf{37.3}   & \textbf{55.0} & \textbf{40.9} & \textbf{21.5}  & \textbf{41.4} & \textbf{53.0}   \\ \hline
                         
\multirow{6}{*}{YOLOv7x~\cite{wang2023yolov7}} & Real-valued & 32-32                & 25.5               & 189.9      & 52.5 & 71.0 & 56.6   & 36.6 & 57.3 & 68.0 \\ \cline{2-11} 
                         & MinMax      & \multirow{3}{*}{8-8} & \multirow{3}{*}{142.6} & \multirow{3}{*}{49.5} & 52.3 & 70.9   & 56.7 & 36.6 & 57.1 & 67.7 \\
                         & Percentile~\cite{li2019fully}  & & &                     & 52.0 & 70.5 & 56.1 & 36.0 & 56.8 & 67.9 \\
                         & \textbf{Q-YOLO}      & & &                      & \textbf{52.4} & \textbf{70.9} & \textbf{56.5} & \textbf{36.2}   & \textbf{57.2} & \textbf{67.8} \\ \cline{2-11} 
                         & Percentile~\cite{li2019fully}  & \multirow{2}{*}{4-4} & \multirow{2}{*}{71.3} & \multirow{2}{*}{25.6} & 36.8 & 55.3 & 40.5 & 21.2 & 41.7 & 49.3 \\
                         & \textbf{Q-YOLO}      & & &                      & \textbf{37.6} & \textbf{57.8}   & \textbf{42.1} & \textbf{23.7} & \textbf{43.8} & \textbf{49.1} \\ \hline
\end{tabular}
\label{exp_main}
\end{table*}

\subsection{Implementation Details}
We randomly selected 1500 training images from the COCO {\tt train2017} dataset \cite{lin2014microsoft} as the calibration data, which served as the foundation for optimizing the model parameters. Additionally, the performance evaluation took place on the COCO {\tt val2017} dataset \cite{lin2014microsoft}, comprising 5000 images. The image size is set to 640x640. 

In our experiments, unless otherwise noted, we employed symmetric channel-wise quantization for weights and asymmetric layer-wise quantization for activations. To ensure a fair and unbiased comparison, we consistently applied the MinMax approach for quantizing weights. The input and output layers of the model are more sensitive to the loss of accuracy. In order to maintain the overall performance of the model, the original accuracy of these layers is usually retained. We also follow this practice.

\subsection{Main results}
We apply our proposed Q-YOLO to quantize YOLOv5s~\cite{yolov5}, YOLOv5m~\cite{yolov5}, YOLOv7~\cite{wang2023yolov7} and YOLOv7x~\cite{wang2023yolov7}, which have an increasing number of parameters.The results of the full-precision model, as well as the 8-bit and 4-bit quantized models using MinMax, Percentile, and Q-YOLO methods, are all presented in Table.\,\ref{exp_main}.

Table.\,\ref{exp_main} lists the comparison of several quantization approaches and detection methods in computing complexity, storage cost. Our Q-YOLO significantly accelerates computation and reduces storage requirements for various YOLO detectors. Similarly, in terms of detection accuracy, when using Q-YOLO to quantize the YOLOv5 series models to 8 bits, there is virtually no decline in the average precision (AP) value compared to the full-precision model. As the number of model parameters increases dramatically, quantizing the YOLOv7 series models to 8 bits results in an extremely slight decrease in accuracy. When quantizing models to 4 bits, the accuracy experiences a significant loss due to the reduced expressiveness of 4-bit integer representation. Particularly, when using the MinMax quantization method, the model loses all its accuracy; whereas the Percentile method, which roughly truncates 99.99\% of the extreme values, fails to bring notable improvement. Differently, Q-YOLO successfully identifies a more appropriate scale for quantization, resulting in a considerable enhancement compared to conventional Post-Training Quantization (PTQ) methods.

\begin{table*}[!t]
\centering
\caption{A comparison of Symmetrical Analysis of Activation Value Quantization. \textbf{\textit{Asymmetric}} indicates the use of an asymmetric activation value quantization scheme, while \textbf{\textit{Symmetric}} refers to the symmetric quantization of activation values.}
\footnotesize
\begin{tabular}{ccccccccccc}
\hline
models                   & Bits                 & Symmetry   & AP   & AP$_{50}$  & AP$_{75}$  & AP$_{s}$  & AP$_{m}$  & AP$_{l}$  \\ \hline
\multirow{5}{*}{YOLOv5s~\cite{yolov5}} & Real-valued          &    -       & 37.4 & 57.1 & 40.1 & 21.6 & 42.3 & 48.9 \\ \cline{2-9} 
                         & \multirow{2}{*}{6-6} & \textit{Asymmetric} & 35.9 & 55.7 & 38.3 & 20.4 & 41.0 & 47.6 \\
                         &                      & \textit{Symmetric}  & 34.4 & 53.9 & 37.0 & 19.3 & 39.8 & 45.0 \\ \cline{2-9} 
                         & \multirow{2}{*}{4-4} & \textit{Asymmetric} & 14.0 & 26.2 & 13.5 & 7.9  & 17.6 & 19.0 \\
                         &                      & \textit{Symmetric}  & 2.7  & 5.9  & 2.2  & 1.3  & 4.2  & 4.6  \\ \hline
\multirow{5}{*}{YOLOv5m~\cite{yolov5}} & Real-valued          &    -       & 45.1 & 64.1 & 49.0 & 28.1 & 50.6 & 57.8 \\ \cline{2-9} 
                         & \multirow{2}{*}{6-6} & \textit{Asymmetric} & 44.0 & 63.1 & 47.7 & 28   & 49.9 & 56.8 \\
                         &                      & \textit{Symmetric}  & 42.4 & 61.1 & 46.0 & 25.3 & 48.3 & 55.9 \\ \cline{2-9} 
                         & \multirow{2}{*}{4-4} & \textit{Asymmetric} & 28.8 & 46.0 & 30.5 & 15.4 & 33.8 & 38.7 \\
                         &                      & \textit{Symmetric}  & 11.3 & 24.8 & 8.6  & 7.5  & 15.2 & 14.5 \\ \hline
\end{tabular}
\label{exp_sys}
\end{table*}

\subsection{Ablation Study}

\subsubsection{Symmetry in Activation Quantization.}
Nowadays, quantization schemes are often subject to hardware limitations; for instance, NVIDIA\cite{nvidia} only supports symmetric quantization, as it is more inference-speed friendly. Therefore, discussing the symmetry in activation value quantization is meaningful. Table.\,\ref{exp_sys} presents a comparison of results using Q-YOLO for symmetric and asymmetric quantization, with the latter exhibiting higher accuracy. The range of negative activation values lies between 0 and -0.2785, while the range of positive activation values exceeds that of the negative ones. If we force equal integer expression bit numbers on both positive and negative sides, the accuracy will naturally decrease. Moreover, this decline becomes more pronounced as the quantization bit number decreases.

\begin{table*}[!t]
\centering
\caption{A comparison of Quantization type. The term \textbf{\textit{only weights}} signifies that only the weights are quantized, \textbf{\textit{only activation}} indicates that only the activation values are quantized, and \textbf{\textit{activation+weights}} represents the quantization of both activation values and weights.}
\footnotesize
\begin{tabular}{ccccccccccc}
\hline
models                   & Bits                 & Quantization type  & AP   & AP$_{50}$  & AP$_{75}$  & AP$_{s}$  & AP$_{m}$  & AP$_{l}$  \\ \hline
\multirow{7}{*}{YOLOv5s\cite{yolov5}} & Real-valued          & -                  & 37.4        & 57.1 & 40.1  & 21.6 & 42.3 & 48.9 \\ \cline{2-9} 
                         & {6-32} & \textit{only weights}       & 36.7(-0.7)  & 56.6 & 39.3  & 20.9 & 41.4 & 48.4 \\
                         & {32-6} & \textit{only activation}    & 36.6(-0.8)  & 56.2 & 39.3  & 21.0 & 42.0 & 47.9 \\
                         & {6-6} & \textit{weights+activation} & 35.9        & 55.7 & 38.3  & 20.4 & 41.0 & 47.6 \\ \cline{2-9} 
                         & {4-32} & \textit{only weights}       & 19.6(-16.3) & 35.6 & 19.3  & 11.3 & 22.5 & 25.7 \\
                         & {32-4} & \textit{only activation}    & 30.6(-5.3)  & 49.1 & 32.6  & 17.0 & 36.7 & 40.7 \\
                         & {4-4} & \textit{weights+activation} & 14.0        & 26.2 & 13.5  & 7.9  & 17.6 & 19   \\ \hline
\multirow{7}{*}{YOLOv5m\cite{yolov5}} & Real-valued          & -                  & 45.1        & 64.1 & 49.0  & 28.1 & 50.6 & 57.8 \\ \cline{2-9} 
                         & {6-32} & \textit{only weights}       & 44.7(-0.4)  & 63.9 & 48.6  & 28.0 & 50.3 & 57.3 \\
                         & {32-6} & \textit{only activation}    & 44.3(-0.8)  & 63.4 & 48.1  & 28.4 & 50.3 & 57.2 \\
                         & {6-6}  & \textit{weights+activation} & 44          & 63.1 & 47.7  & 28.0 & 49.9 & 56.8 \\ \cline{2-9} 
                         & {4-32} & \textit{only weights}       & 34.6(-9.4)  & 54.0 & 37.3  & 20.0 & 39.2 & 45.3 \\
                         & {32-4} & \textit{only activation}    & 37.7(-6.3)  & 57.3 & 41 .8 & 23.7 & 44.1 & 51.0 \\
                         & {4-4}  & \textit{weights+activation} & 28.8        & 46.0 & 30.5  & 15.4 & 33.8 & 38.7 \\ \hline
\end{tabular}
\label{exp_mode}
\end{table*}

\subsubsection{Quantization Type.}
In Table.\,\ref{exp_mode}, we analyze the impact of different quantization types on the performance of the YOLOv5s and YOLOv5m models, considering three cases: quantizing only the weights (\textit{only weights}), quantizing only the activation values (\textit{only activation}), and quantizing both weights and activation values (\textit{weights+activation}). The results demonstrate that, compared to quantizing the activation values, quantizing the weights consistently induces larger performance degradation. Additionally, the lower the number of bits, the greater the loss incurred by quantization. In YOLO, the weights learned by a neural network essentially represent the knowledge acquired by the network, making the precision of the weights crucial for model performance. In contrast, activation values serve as intermediate representations of input data propagating through the network, and can tolerate some degree of quantization error to a certain extent.

\subsection{Inference speed}
To practically verify the acceleration benefits brought about by our quantization scheme, we conducted inference speed tests on both GPU and CPU platforms. For the GPU, we selected the commonly used desktop GPU {\tt NVIDIA RTX 4090}~\cite{nvidia} and the {\tt NVIDIA Tesla T4}~\cite{nvidia} , often used in computing centers for inference tasks. Due to our limited CPU resources, we only tested Intel products, the {\tt i7-12700H} and {\tt i9-10900}, both of which have x86 architecture. For deployment tools, we chose TensorRT~\cite{tensorrt} and OpenVINO~\cite{openvino}. The entire process involved converting the weights from the torch framework into an ONNX model with QDQ nodes and then deploying them onto specific inference frameworks. The inference mode was set to single-image serial inference, with an image size of 640x640. As most current inference frameworks only support symmetric quantization and 8-bit quantization, we had to choose a symmetric 8-bit quantization scheme, which resulted in an extremely small decrease in accuracy compared to asymmetric schemes. As shown in Table.\,\ref{exp_speed}, the acceleration is extremely significant, especially for the larger YOLOv7 model, wherein the speedup ratio when using a GPU even exceeded \textbf{3$\times$} compared to the full-precision model. This demonstrates that applying quantization in real-time detectors can bring about a remarkable acceleration.

\begin{table*}[!t]
\centering
\caption{The inference speed of the quantized model is essential. The quantization scheme adopts uniform quantization, with single-image inference mode and an image size of 640*640. TensorRT~\cite{tensorrt}is selected as the GPU inference library, while OpenVINO~\cite{openvino} is chosen for the CPU inference library}

\begin{tabular}{ccccccccccc}
\hline
\multirow{2}{*}{models}  & \multirow{2}{*}{Bits}  &\multirow{2}{*}{AP} & \multicolumn{2}{c}{GPU speed /\textit{ms}} & \multicolumn{2}{c}{Intel CPU speed /\textit{ms}} \\ \cline{4-7} 
                         &                        &              &RTX 4090        & Tesla T4        & i7-12700H(x86)   & i9-10900(x86)  \\ \hline
\multirow{2}{*}{YOLOv5s} & 32-32                  &37.4        &4.9             & 7.1             & 48.7             & 38.7           \\ \cline{2-7} 
                         & 8-8                    &37.3 & 3.0              & 4.5            & 33.6             & 23.4           \\ \hline
\multirow{2}{*}{YOLOv7}  & 32-32                  &50.8 &16.8            & 22.4            & 269.8            & 307.8          \\ \cline{2-7} 
                         & 8-8                    &50.6 &5.4             & 7.8             & 120.4            & 145.2          \\ \hline
\end{tabular}
\label{exp_speed}
\end{table*}

\section{Conclusions}
Real-time object detection is crucial in various computer vision applications. However, deploying object detectors on resource-constrained platforms poses challenges due to high computational and memory requirements. This paper introduces Q-YOLO, a highly efficient one-stage detector built using a low-bit quantization method to address the performance degradation caused by activation distribution imbalance in traditional quantized YOLO models. Q-YOLO employs a fully end-to-end Post-Training Quantization (PTQ) pipeline with a well-designed Unilateral Histogram-based (UH) activation quantization scheme. Extensive experiments conducted on the COCO dataset demonstrate the effectiveness of Q-YOLO. It outperforms other PTQ methods while achieving a favorable balance between accuracy and computational cost. This research significantly contributes to advancing the efficient deployment of object detection models on resource-limited edge devices, enabling real-time detection with reduced computational and memory requirements. 
%
%
%
\bibliographystyle{splncs04}
\bibliography{references}
\end{document}